\documentclass[conference]{IEEEtran}
\IEEEoverridecommandlockouts
\usepackage{cite}
\usepackage{amsmath,amssymb,amsfonts}
\usepackage{algorithmic}
\usepackage{graphicx}
\usepackage{textcomp}
\usepackage{xcolor}
\usepackage{caption}
\usepackage{subcaption}
\def\BibTeX{{\rm B\kern-.05em{\sc i\kern-.025em b}\kern-.08em
    T\kern-.1667em\lower.7ex\hbox{E}\kern-.125emX}}
\begin{document}

\title{The Diabetic Buddy: A Diet Regulator and Tracking System for Diabetics\\}

\author{\IEEEauthorblockN{Muhammad Usman}
\IEEEauthorblockA{\textit{Division of Information and Computing Technology} \\
\textit{College of Science and Engineering}\\
\textit{Hamad Bin Khalifa University}\\
Doha, Qatar \\
musman@hbku.edu.qa}
\and
\IEEEauthorblockN{Kashif Ahmad}
\IEEEauthorblockA{\textit{Division of Information and Computing Technology} \\
\textit{College of Science and Engineering}\\
\textit{Hamad Bin Khalifa University}\\
Doha, Qatar \\
kahmad@hbku.edu.qa}
\and
\IEEEauthorblockN{Amir Sohail}
\IEEEauthorblockA{\textit{Department of Information Engineering} \\
\textit{and Computer Science} \\
\textit{University of Trento}\\
Trento, Italy \\
amir.sohail@studenti.unitn.it}
\and
\IEEEauthorblockN{Marwa Qaraqe}
\IEEEauthorblockA{\textit{Division of Information and Computing Technology} \\
\textit{College of Science and Engineering}\\
\textit{Hamad Bin Khalifa University}\\
Doha, Qatar \\
mqaraqe@hbku.edu.qa}
}

\maketitle

\begin{abstract}
The prevalence of Diabetes mellitus (DM) in the Middle East is exceptionally high as compared to the rest of the world. In fact, the prevalence of diabetes in the Middle East is 17-20\%, which is well above the global average of 8-9\%. Research has shown that food intake has strong connections with the blood glucose levels of a patient. In this regard, there is a need to build automatic tools to monitor the blood glucose levels of diabetics and their daily food intake. This paper presents an automatic way of tracking continuous glucose and food intake of diabetics using off-the-shelf sensors and machine learning, respectively. Our system not only helps diabetics to track their daily food intake but also assists doctors to analyze the impact of the food in-take on blood glucose in real-time. For food recognition, we collected a large-scale Middle-Eastern food dataset and proposed a fusion-based framework incorporating several existing pre-trained deep models for Middle-Eastern food recognition.  

\end{abstract}

\begin{IEEEkeywords}
Diabetes management, food recognition, continuous glucose monitoring, middle-eastern food
\end{IEEEkeywords}

\section{Introduction}
Diabetes mellitus (DM) is a disorder that arises from a dysfunction of the glucoregulatory system. Over 463 million people worldwide are diabetic and this number is projected to jump to 700 million by 2040 \cite{atlas2015international}. With its high prevalence worldwide, diabetes is becoming one of the fastest-growing health challenges of the 21st century and is costing the healthcare system and wider global economy an estimated 825 billion US dollars annually \cite{fagherazzi2019digital}. 

Diabetes requires constant management and, in addition to medication, self-management of
diabetes is vital in preventing acute complications and minimizing the risk of long-term complications. The entire paradigm of diabetes management has been transformed in the past decade due to the development of new technologies such as continuous glucose monitoring (CGM) devices. Diabetes management has become an ecosystem wherein food is an important element that directly affects blood glucose levels in the human body. A careful meal plan has the potential to not only regulate the blood glucose level but also to keep it under the desired range \cite{seligman2007food}. 

In fact, what a person eats, when s/he eats, and the amount s/he eats affect the blood glucose levels in the body. The normal range of blood glucose in our body has a direct relation with the aforementioned questions. A blood glucose level considered high at eight hours of fasting may be considered normal just after eating or even two hours after eating. In this way, the blood glucose level strongly depends on the type of food and time it was eaten. Table \ref{table:glucoselevel} highlights the relation between food intake and blood glucose levels for non-diabetics, pre-diabetics, and diabetics. 

The current diabetes management system requires patients to track their meals via a paper-based method. Although the process of CGM can be automated using off-the-shelf sensors, such as FreeStyle Libre sensors \cite{rodbard2016continuous}, the process of automating food tracking still lacks in the literature. Even though there have been some applications aimed at food tracking, none of these applications can precisely correlate meal time and type with CGM. CGM application, such as the one from FreeStyle Libre, allows patients to manually write the meal-type and time on the applications but this is not an efficient way as a patient may forget to write in time. On the other hand, food tracking applications have no option for CGM integration. Moreover, these applications are mainly designed for Western Food, not local Middle-Eastern food. So this paper tries to address two important gaps in the literature, (i) Mapping of food in-take to CGM graph for better understanding of meal impact on blood glucose levels and (ii) development of deep learning algorithms to correctly recognize the Middle-Easters cuisines.

This work presents continuous glucose and food tracker (CGFT) for diabetics in the Middle-East region having the ability to track real-time blood glucose levels along with the food intake of the patient. The main highlights of the proposed system are given below.

\begin{itemize}
    \item Being an integral part of the proposed application, the application is equipped with a novel fusion-based middle eastern meal recognizer covering most of the local Arab cuisines. 
   % \item We also provide detailed analysis of different feature extraction algorithms is performed on (i) handcrafted visual features and (ii) deep features
   % \item As the results from deep features outperforms the handcrafted features, they are translated into a mobile application used by diabetics in the Middle-East to capture their daily meal in-take
    \item An interface to collect data from the FreeStyle Libre Sensor is developed, which connects the mobile application with the sensor over Bluetooth connection.
    \item A web-platform is developed for the doctors and the patients, which stores the data collected from the mobile application. 
    \item A mapping of CGM and the food-intake is provided to see the real-time impact of the food on blood glucose.
\end{itemize}

For the demonstration purposes, we will give a live demo of the system where it will be demonstrated how our system is able to help diabetics by tracking continuous glucose and food intake using off-the-shelf sensors and ML. 

%The rest of the paper is organized as follows. Section ..
% \usepackage{booktabs}
\begin{table}[]
\caption{Effect of food on blood glucose levels of non-diabetics, pre-diabetics and diabetics \cite{Mayoclicin}.}
\label{table:glucoselevel}
\centering
\scalebox{0.9}{
%\begin{tabular}{@{}p{1.7cm}p{1.1cm}p{1.8cm}p{2.3cm}@{}}
\begin{tabular}{p{1.7cm}p{1.1cm}p{1.8cm}p{2.3cm}}
  & \textbf{Fasting (mg/dl)} & \textbf{After Eating (mg/dl)} & \textbf{2 Hours After Eating (mg/dl)} \\ \hline \hline
\textbf{Non-Diabetic}       & 80-100  & 170-200 & 120-140 \\ 
\textbf{Pre-Diabetic} & 101-125 & 190-230 & 140-160 \\ 
\textbf{Diabetic}     & 126+    & 220-300 & 200+    \\ \hline
\end{tabular}
}
\end{table}

\section{Proposed System}
\begin{figure}[]
\centering
\includegraphics[width=0.450\textwidth]{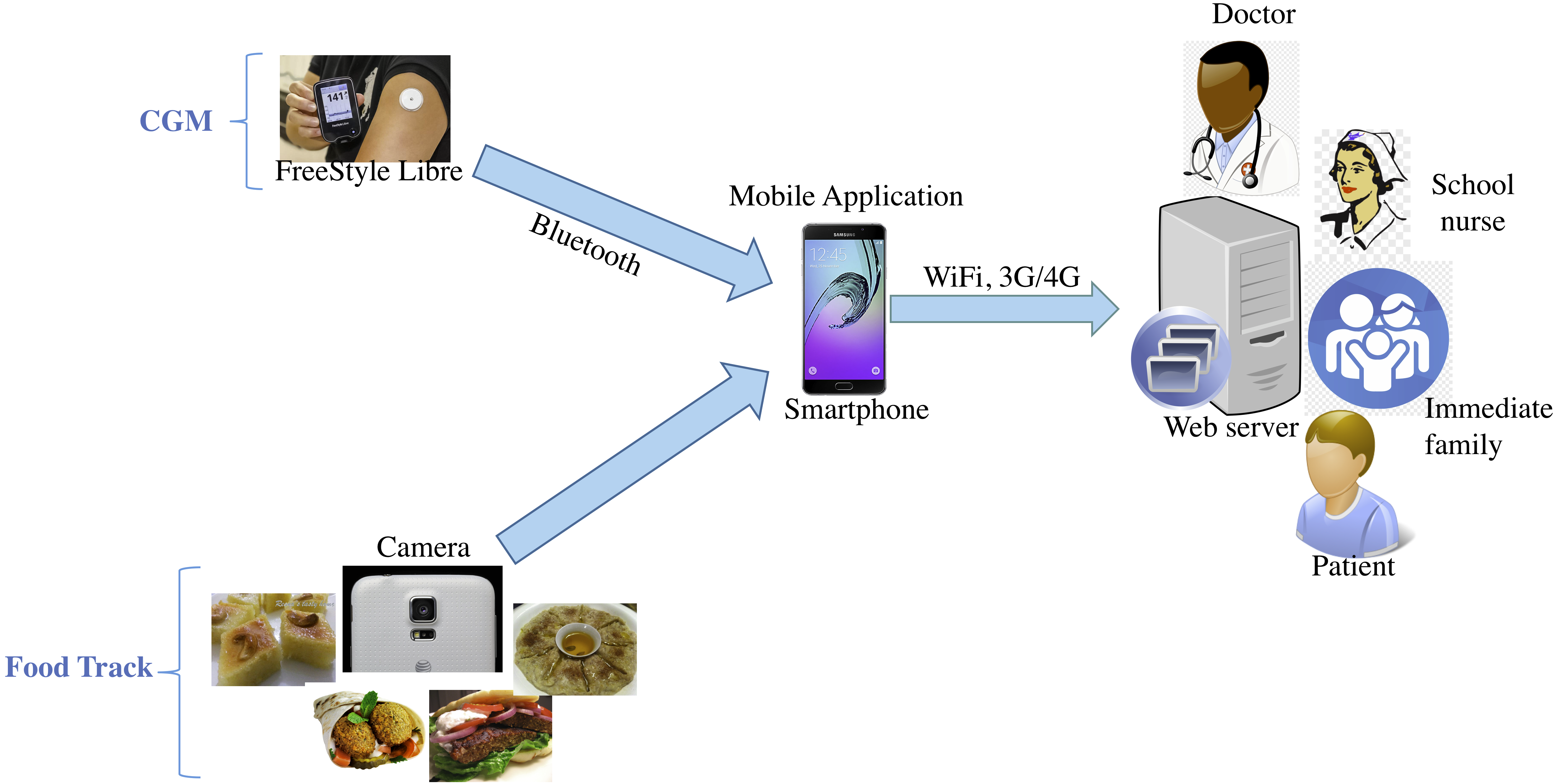}
\caption{Systems architecture of the proposed system.}
\label{fig:sys_architecture}
\end{figure}

The architecture of the proposed system is presented in Figure \ref{fig:sys_architecture}. There are two main data sources in the application. one is for continuous glucose readings from the FreeStyle Libre sensor worn by the patient. The data is stored in mobile applications and the web-platform and plotted as a CGM graph. %No data-analytics is performed on this data. 
The second source of data is the smartphone's camera, which a patient uses to take the picture of his meal. This data is stored in the mobile application and the web-platform. Using data analytics, the food item is identified and mapped in the CGM graph.

%%%%%%%%%%%%%%%%%%%%%Block Diagram of App%%%%%%%%%%%%%%%%%%%%%%%%%%
%%%%%%%%%%%%%%%%%%%%%%%%%%%%%%%%%%%%%%%%%%%%%%%%%%%%%%%%%%%%%%%%%%%

The details of the proposed system in terms of data collection, storage, and analytics are given below. 

\subsection{CGM Data}
\begin{figure}[]
\centering
\includegraphics[width=0.20\textwidth]{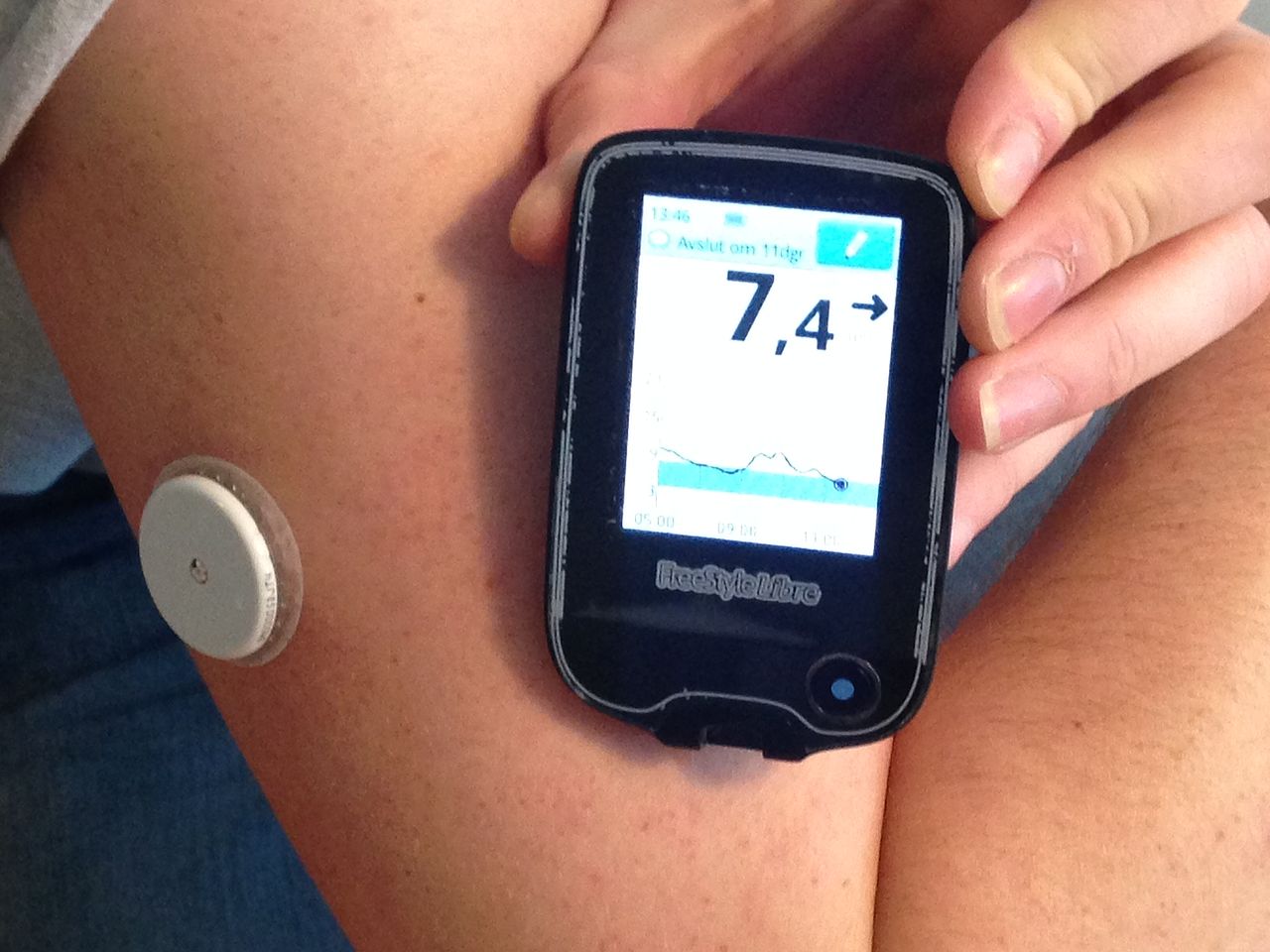}
\caption{FreeStyle Libre Sensor with reader.}
\label{fig:freestyle}
\end{figure}

\begin{figure}[]
\centering
\includegraphics[width=0.15\textwidth]{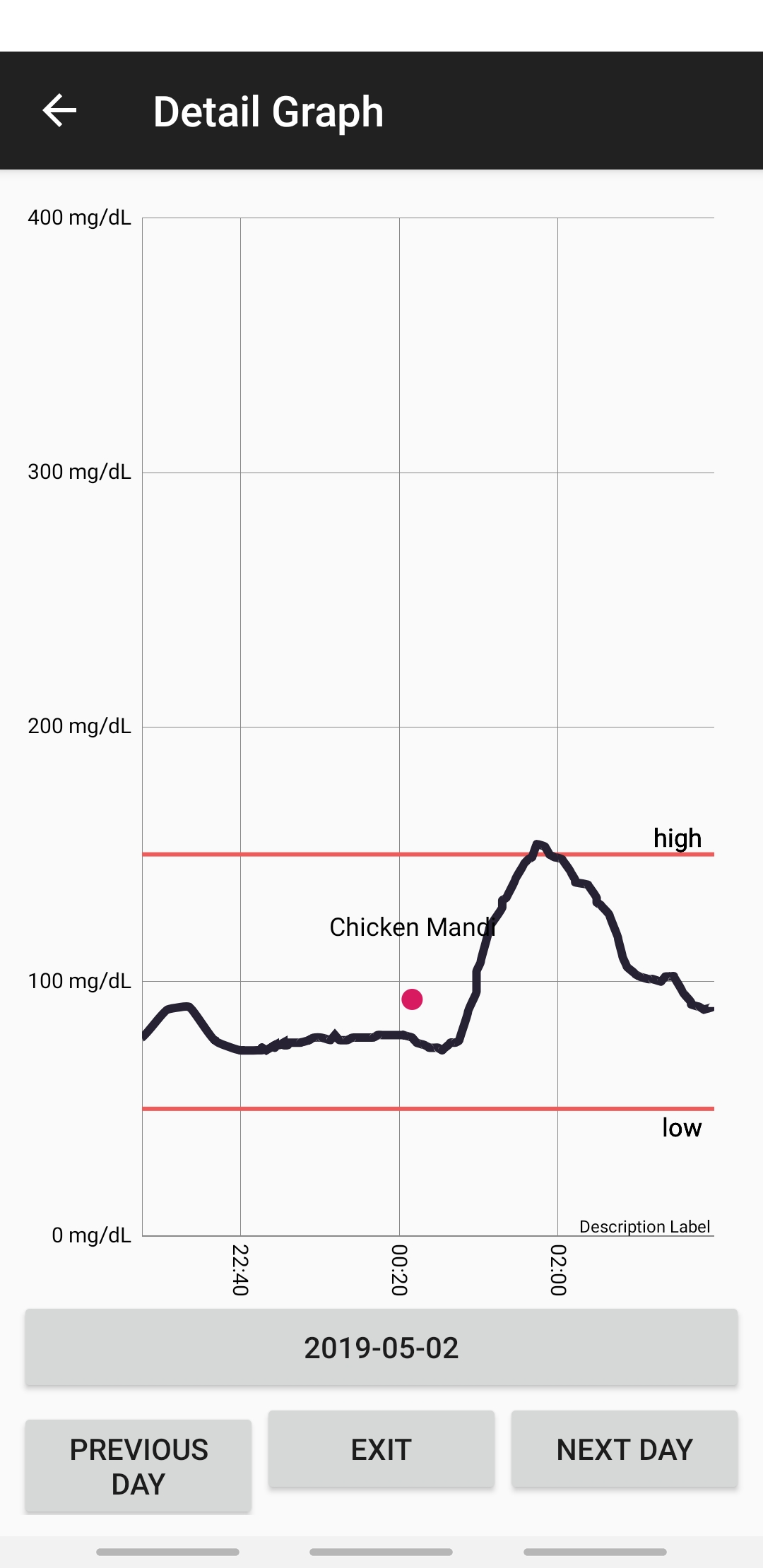}
\caption{CGM plot on the mobile application.}
\label{fig:cgm}
\end{figure}

For continuous glucose monitoring, we rely on FreeStyle Libre Sensors. These are invasive sensors that take a reading every 15 minutes. An illustration of these sensors is given in Figure \ref{fig:freestyle}. The sensor can store eight hours of data. Intrinsically, the data from these sensors can only be read by Near Field Communication (NFC) of a smartphone or the reader shown in Figure \ref{fig:freestyle}. However, this is not convenient for the patient to tap the reader/smartphone time and again. In order to solve this problem, we use an NFC-to-Bluetooth converter named NightRider BluCon \footnote{https://www.ambrosiasys.com} to make this data transmission seamless to the patient. A reading is sent to the smartphone every five minutes, which is plotted in terms of a time graph (CGM plot). A snapshot of this graph on the mobile application is shown in Figure \ref{fig:cgm}. %It is important to note that details of the communication  protocols to connect the BluCon with the developed mobile application are out of the scope of this paper.

%%%%%%%%%%%%%%%%%%%%%Block Diagram of App%%%%%%%%%%%%%%%%%%%%%%%%%%
%%%%%%%%%%%%%%%%%%%%%%%%%%%%%%%%%%%%%%%%%%%%%%%%%%%%%%%%%%%%%%%%%%%

%%%%%%%%%%%%%%%%%%%%%Block Diagram of App%%%%%%%%%%%%%%%%%%%%%%%%%%
%%%%%%%%%%%%%%%%%%%%%%%%%%%%%%%%%%%%%%%%%%%%%%%%%%%%%%%%%%%%%%%%%%%

\subsection{Food Data}
\begin{figure}[]
\centering
\includegraphics[width=0.15\textwidth]{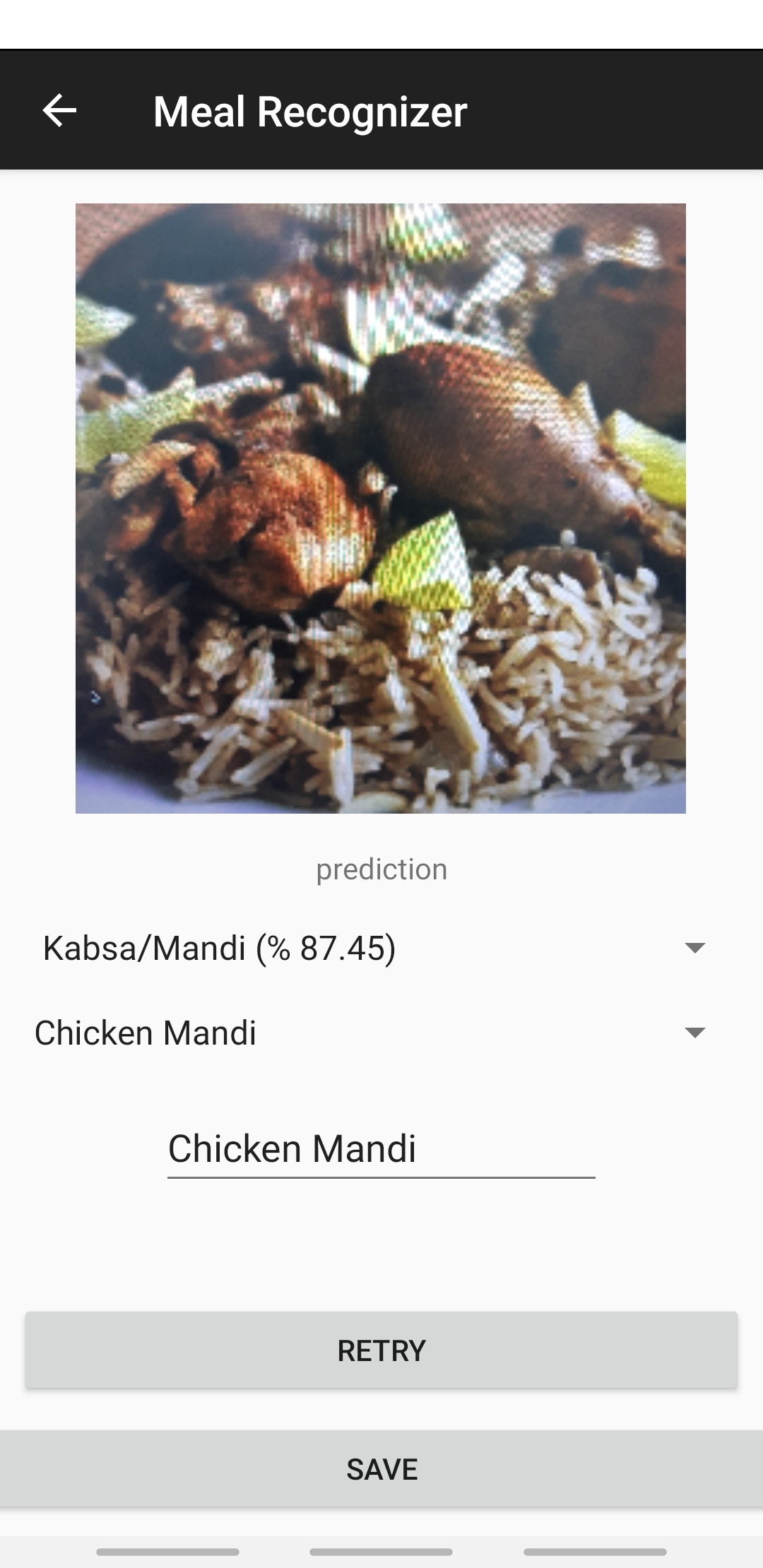}
\caption{Meal recognizer in  the mobile application.}
\label{fig:app}
\end{figure}

Before every meal, the patient has to take the picture of his meal. A food recognition framework identifies the food item in the picture. The identified food item is then plotted on the CGM graph. For instance, chicken mandi is identified in Figure \ref{fig:cgm} that the patient ate at around mid-night. A mobile snapshot of the meal recognizer is presented in Figure \ref{fig:app}. The details of the food detection algorithm and the collected dataset are provided in subsequent sub-sections. 

%%%%%%%%%%%%%%%%%%%%%Block Diagram of App%%%%%%%%%%%%%%%%%%%%%%%%%%
%%%%%%%%%%%%%%%%%%%%%%%%%%%%%%%%%%%%%%%%%%%%%%%%%%%%%%%%%%%%%%%%%%%

\subsection{Dataset}
For the proposed application, we collected a large-scale dataset covering images from 38 different categories of Middle-Eastern food. One of the biggest challenges was the availability of traditional Middle-Eastern food images. In order to collect a sufficient number of images, we crawled online sources, such as Google and Flicker search engines to collect 7530 images from 38 different categories. However, some of the food categories resembled a lot in terms of their appearance and ingredients. For instance, mandi and kabsa are not only visually similar but their ingredients are also similar. Indeed, it is very difficult for a human observer to differentiate among them. Based on this observation, we merged some of the food categories and added a drop-down menu for patients to select among merged categories. %All the experiments are executed on the second version of the dataset wherein various similar categories are merged. 

%%%%%%%%%%%%%%%%%%%%%Block Diagram of App%%%%%%%%%%%%%%%%%%%%%%%%%%
\begin{figure}[]
\centering
\includegraphics[width=0.42\textwidth]{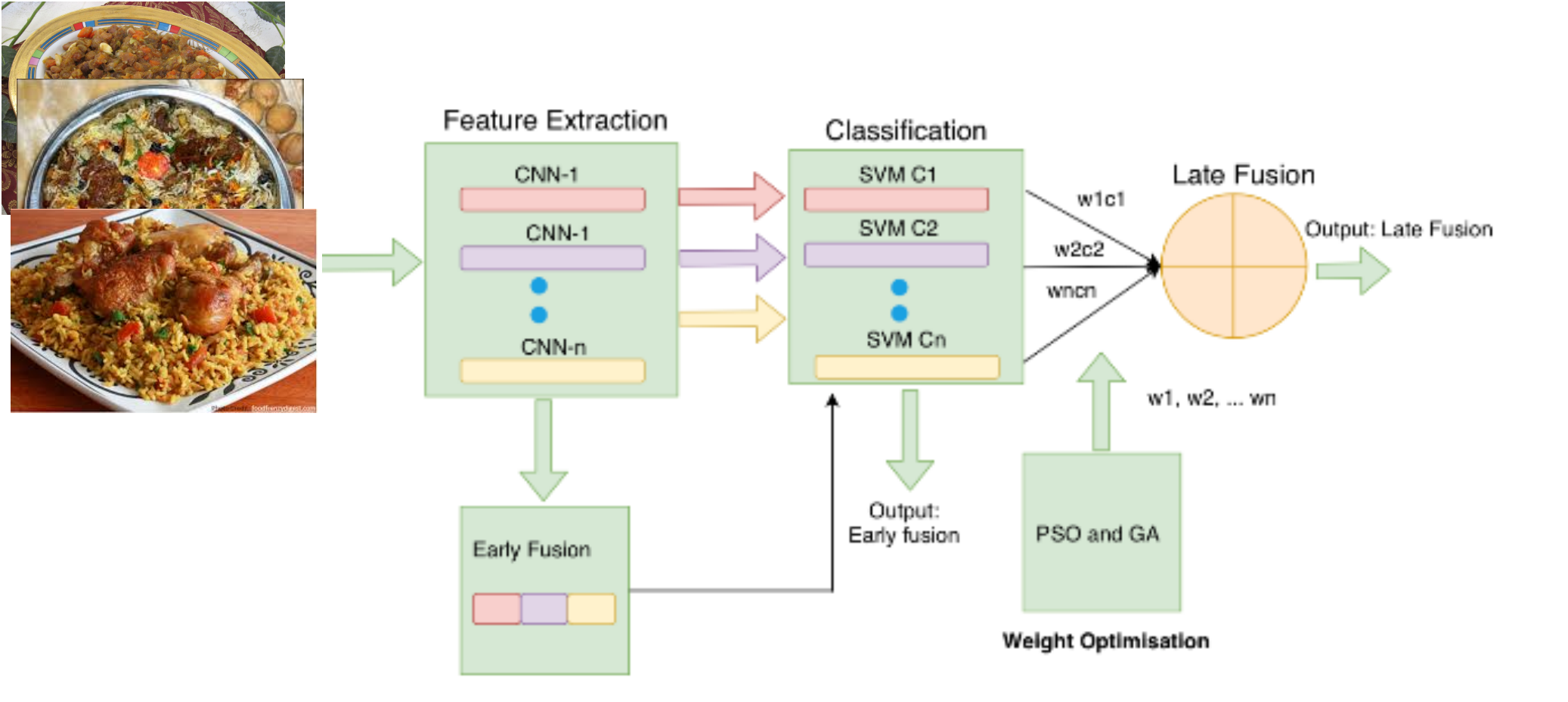}
\caption{Block diagram of the Food Recognizer.}
\label{fig:Block_Diagram_methodology_food}
\end{figure}
%%%%%%%%%%%%%%%%%%%%%%%%%%%%%%%%%%%%%%%%%%%%%%%%%%%%%%%%%%%%%%%%%%%

%%%%%%%%%%%%%%%%%%%%%Block Diagram of App%%%%%%%%%%%%%%%%%%%%%%%%%%
\begin{figure}[]
\centering
\includegraphics[width=0.40\textwidth]{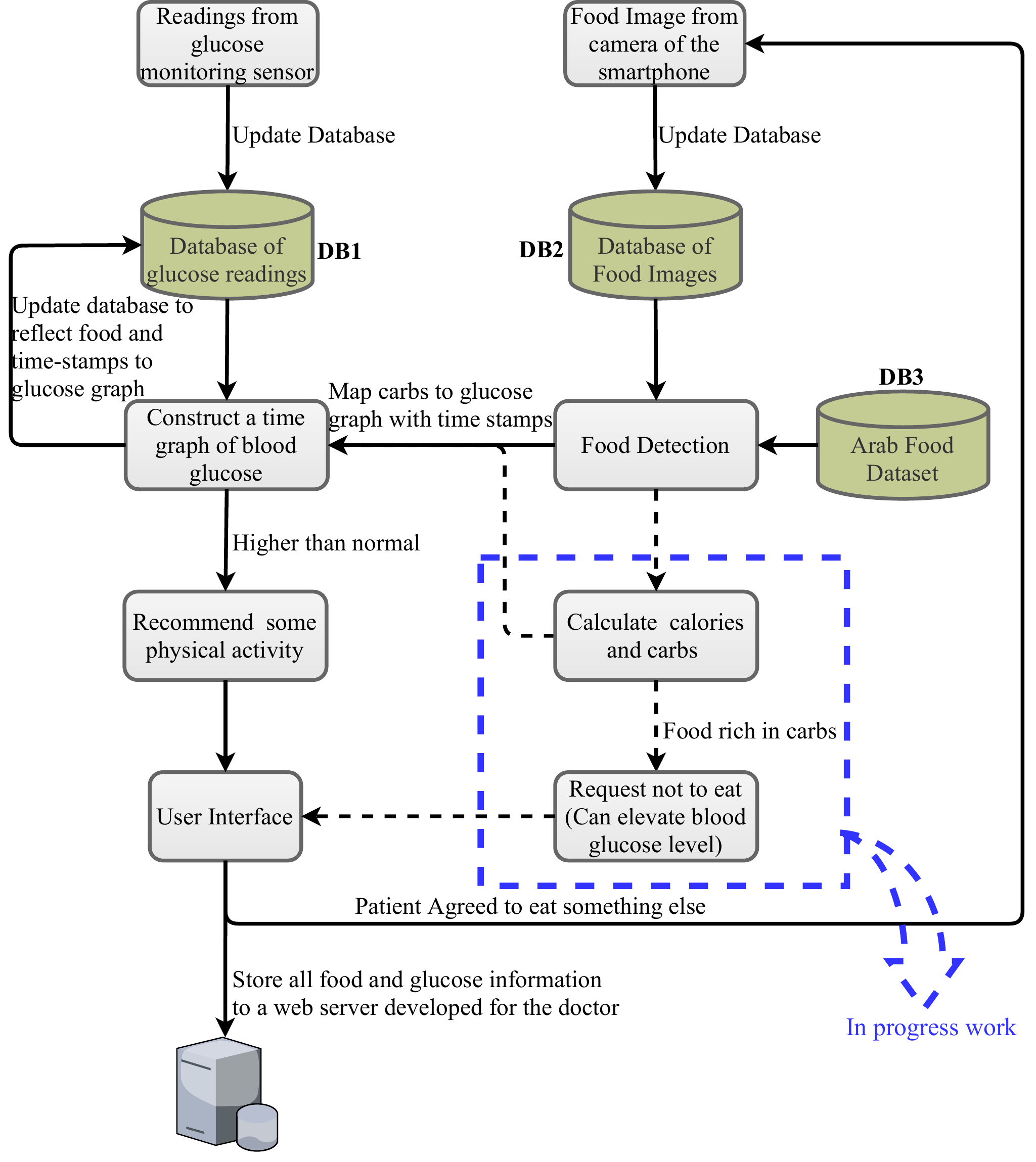}
\caption{Block diagram of mobile application}
\label{fig:Block_Diagram_App}
\end{figure}
%%%%%%%%%%%%%%%%%%%%%%%%%%%%%%%%%%%%%%%%%%%%%%%%%%%%%%%%%%%%%%%%%%%

%%%%%%%%%%%%%%%%%%%%%%%%%%%%%%%%%%%%%%%%%%%%%%%%%%%%%%%%%%%%%%%%%%%%%%%%%%%%%%%
\begin{table}[!h]
\caption{Experimental results of the middle eastern meal recognizer in terms of precision, recall, F-score and accuracy.}
\label{table:fused_results_v2}
\centering
%\scalebox{0.8}{
\begin{tabular}{|c|c|c|c|c|}
\hline
\textbf{Method} & \textbf{Avg. Pre.} & \textbf{Avg. Rec.} & \textbf{Avg. F-Score} &\textbf{Acc.} \\ \hline
PSO based Fusion  & 70.89 & 79.71 & 75.04 & 80.82 \\ \hline
GA based Fusion  & 69.83 & 77.30 & 73.37 & 79.61 \\ \hline
Early Fusion & 66.08 & 73.10  & 69.41 & 76.53 \\ \hline
Equal weights  & 70.74  & 76.74 & 73.61 & 78.89 \\ \hline
%Singla et al. \cite{singla2016food} & 65.40 & 71.90 & 68.49 & 75.49 \\ \hline
%Aguilar et al. \cite{aguilar2017food} & 68.71 & 75.53 & 71.95 & 78.73 \\ \hline
\end{tabular}
%}
\end{table}
%%%%%%%%%%%%%%%%%%%%%%%%%%%%%%%%%%%%%%%%%%%%%%%%%%%%%%%%%%%%%%%%%%%%%%%%%%%%%
%********************************************************************

\begin{figure*}
\begin{subfigure}{.45\textwidth}
  \centering
  % include first image
  \includegraphics[width=.8\linewidth]{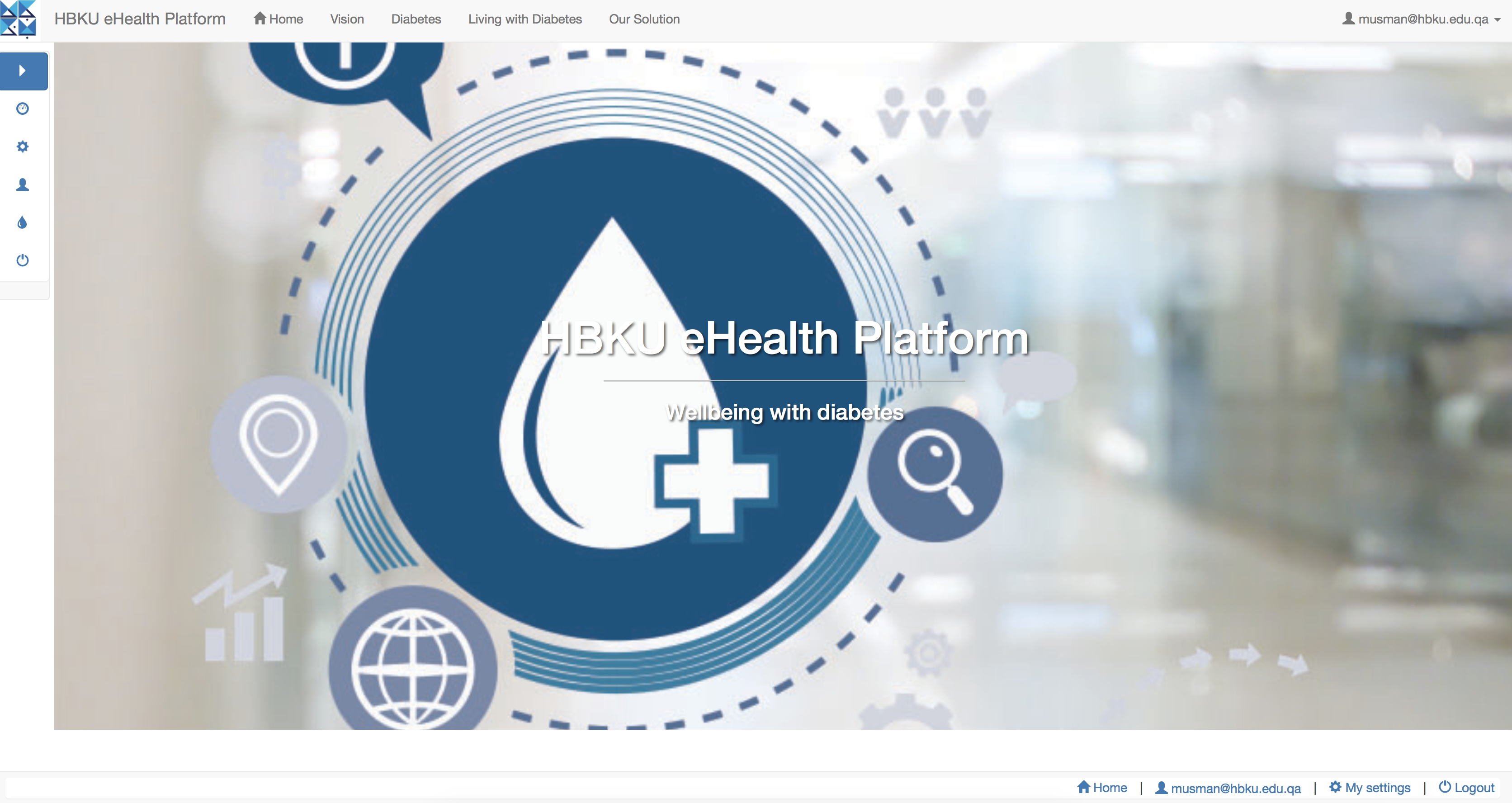}  
  \caption{Home page}
  \label{fig:sub-first}
\end{subfigure}
\begin{subfigure}{.45\textwidth}
  \centering
  % include second image
  \includegraphics[width=.8\linewidth]{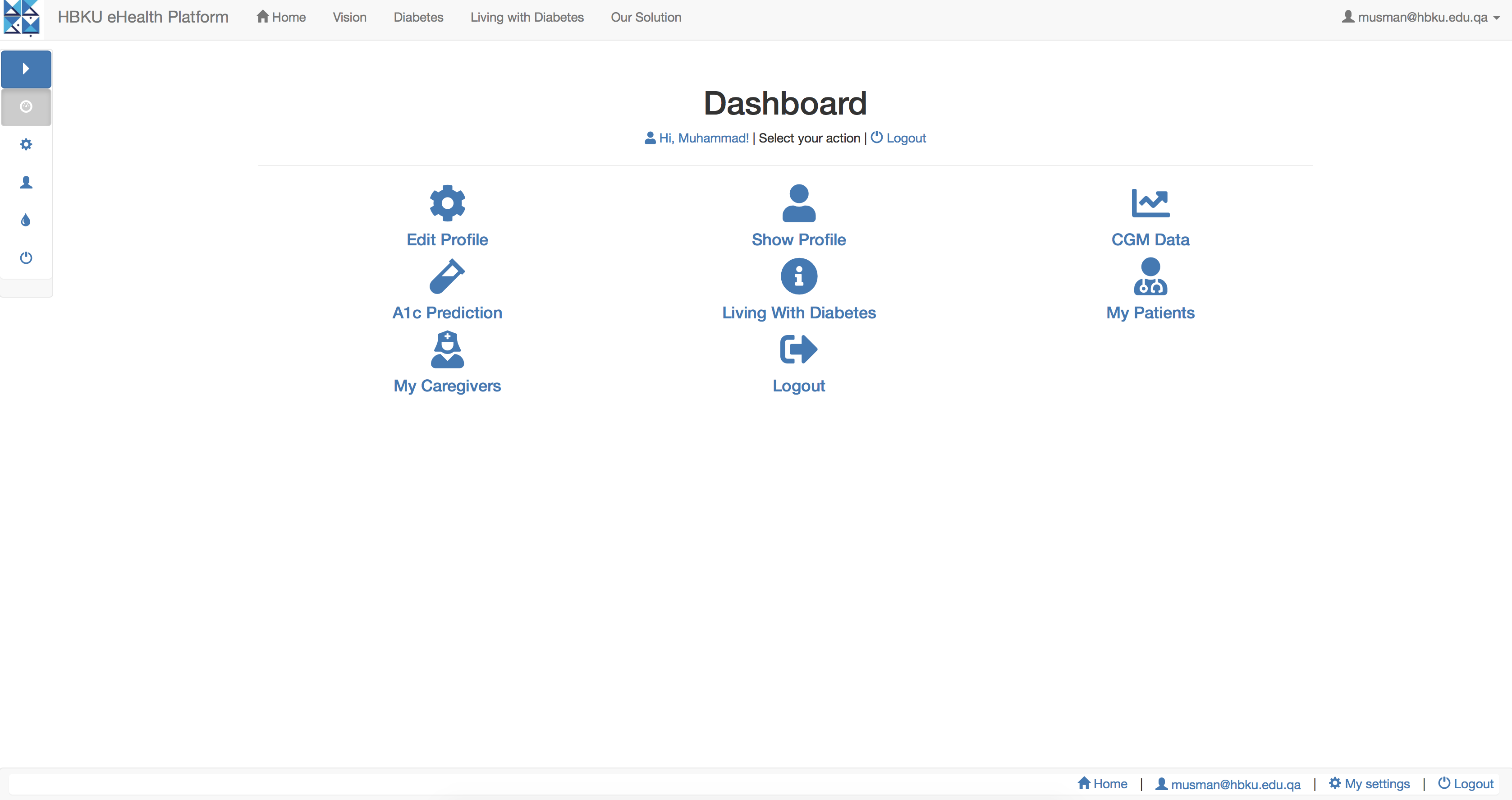}  
  \caption{Dashboard}
  \label{fig:sub-second}
\end{subfigure}

\begin{subfigure}{.45\textwidth}
  \centering
  % include third image
  \includegraphics[width=.8\linewidth]{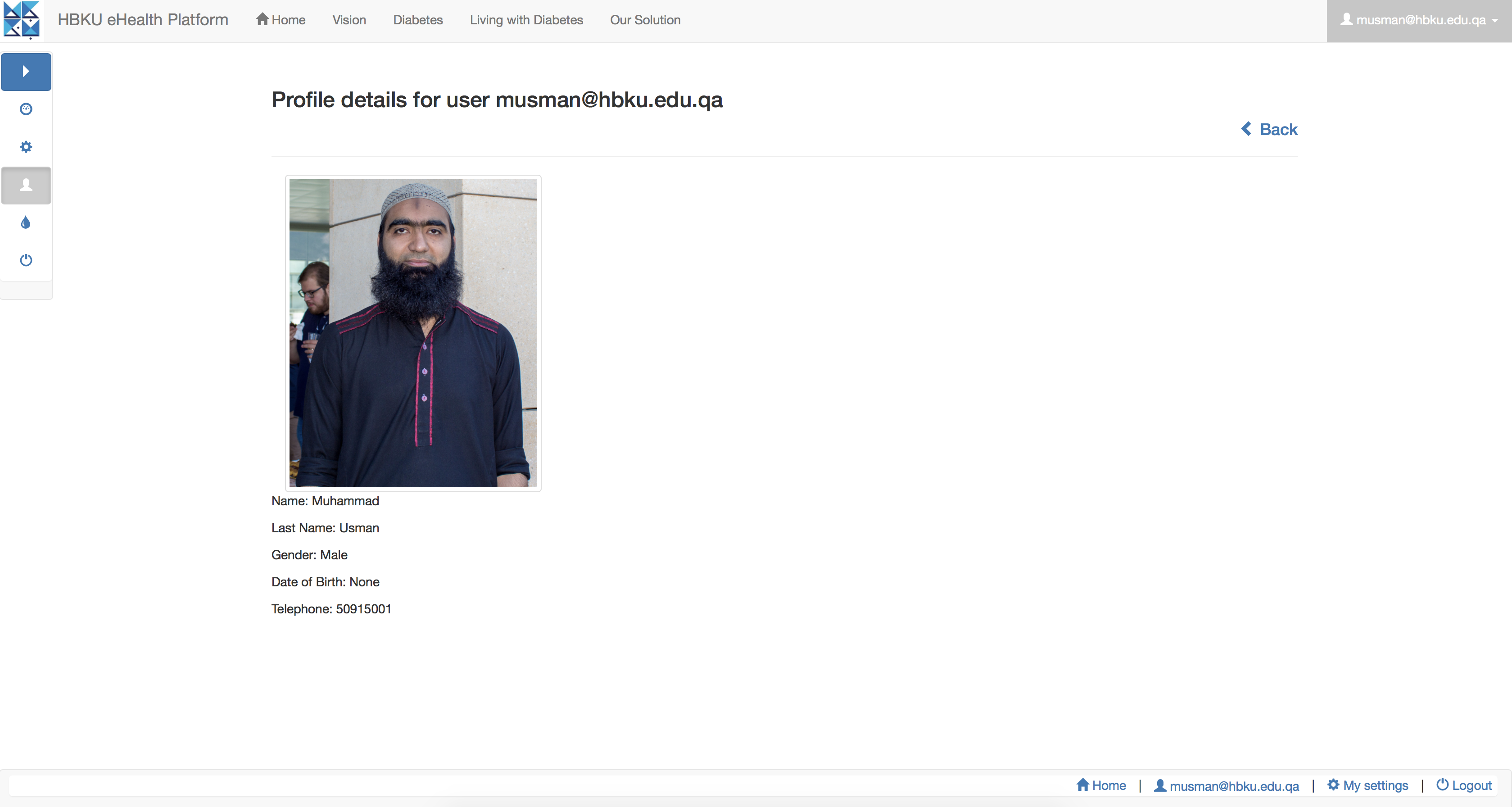}  
  \caption{User profile}
  \label{fig:sub-third}
\end{subfigure}
\begin{subfigure}{.45\textwidth}
  \centering
  % include fourth image
  \includegraphics[width=.7\linewidth]{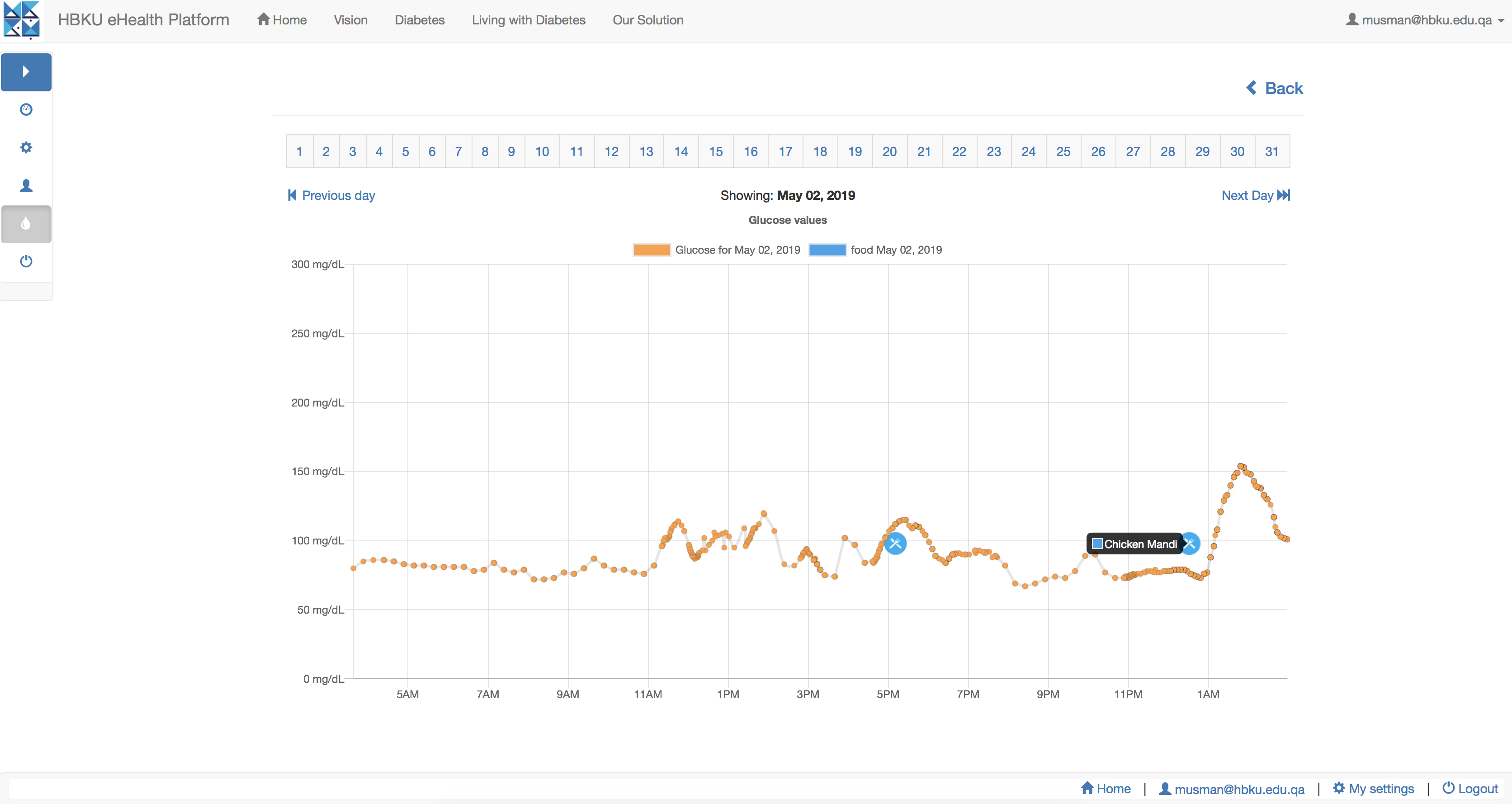}  
  \caption{CGM and meal recognizer}
  \label{fig:sub-fourth}
\end{subfigure}
\caption{Few snapshots of the web-platform}
\label{fig:webplatform}
\end{figure*}
\subsection{Middle-Eastern Meal Recognizer} \label{subsec:mr}
Figure \ref{fig:Block_Diagram_methodology_food} provides block diagram of the methodology adopted for the food/meal recognizer module. As can be seen, the proposed methodology is composed of three main components, namely (i) feature extraction, (ii) classification and (iii) fusion. As a first stept, features are extracted via several existing pre-trained deep models including AlexNet \cite{krizhevsky2012imagenet}, VggNet \cite{simonyan2014very}, GoogleNet \cite{szegedy2015going} and ResNet \cite{he2016deep}. It is important to mention that these models are used as features descriptors only, where features are extracted from the last fully connected layers without any fine-tuning and retraining. After feature extraction, SVMs are trained on the features extracted through each individual model. 

The final part of the methodology is based on the early (simple concatenation of feature vectors) and three different late fusion methods, namely Particle Swarm based Optimization (PSO), Genetic Algorithms (GA), and simple averaging of the scores obtained with the individual models. The fusion of multiple deep models is motivated by the fact that the deep models/architectures respond differently to an image by extracting diverse but complementary features. Combining such diverse but complementary features could result in an improvement in the performance of our middle eastern meal recognizer. Moreover, based on our previous experience, we believe merit-based weights should be assigned to each model during the fusion. The selection of PSO and GA based weight optimization of the weights to be used in the fusion is based on their proven performance in other application domains \cite{ahmadfood,ahmad2018ensemble}. 

Our fitness function for PSO and GA based fusion methods is based on the accumulative error computed on a validation set using Equ. 1 where \textit{$A_{acc}$} is computed using Equ. 2. Here \textit{$p_{n}$} represents the probabilities obtained through \textit{$n^{th}$} model on the validation set while \textit{$x(n)$} represents the value (weight) to be used and optimized for the \textit{$n^{th}$} model.
%%%%%%%%%%%%%%%%%%%%%%%%%%%%%%%%%%%%%%%%%%%%%%
\begin{equation}
error_{acc} = 1-A_{acc}
%	\label{fitness_function}
\end{equation}
%%%%%%%%%%%%%%%%%%%%%%%%%%%%%%%%%%%%%%%%%
\begin{equation}
%\small
%\centering
A_{acc} = x(1)*p_{1}+x(2)*p_{2}+... +x(n)*p_{n} 
\end{equation}
%%%%%%%%%%%%%%%%%%%%%%%%%%%%%%%%%%%%%%%%

Table \ref{table:fused_results_v2} provides the experimental results of the proposed middle eastern meal recognizer in terms of precision, recall, F1-score, and accuracy. As can be seen, merit-based fusion (PSO and GA) have a clear advantage over the simple averaging and early fusion method.

\subsection{Mobile Application}
One of the most important components of our solution is the mobile application that collects data from CGM sensors over Bluetooth and plots it on the time graph. We integrate the meal recognizer discussed in subsection \ref{subsec:mr} into the application. The application recognizes the meal and maps it over the CGM plot as shown in \ref{fig:Block_Diagram_App}. Glucose readings are stored in a database (DB) 1, while every image taken by the patient is stored in DB2. DB3 contains the collected dataset on which the system is trained. The application has the ability to alert the patient in case the blood glucose level is high  and recommend some physical activity. An alert can be sent to the doctor and immediate family members if the glucose level is very high. In addition, the patient can be recommended not to eat high carb food.

\subsection{Web-Platform}
The system also provides a web-platform for the doctors, patients, and immediate family members to closely monitor the CGM and food-intake of the patient. All the information stored in the mobile application is reflected in the web-platform as well. Few snapshots of the developed application are presented in Figure \ref{fig:webplatform}. It is important  to note that CGM values plotted in Figure \ref{fig:sub-fourth} exactly duplicate the ones plotted in Figure \ref{fig:cgm}.

\section{Conclusion}
This paper addresses two important shortcomings of the literature. Firstly, it maps the food intake of a patient on his/er CGM graph. Secondly, the food recognition algorithm is customized for local Middle-Eastern Cuisines. The end-system is a combination of a mobile application and a web-platform that graphically represents the CGM and food-intake for the patients and the doctors, respectively. In the future, we aim to estimate the number of carbs and calories in the meal.

\section*{Acknowledgment}
This work was supported by internal fund of HBKU.

\bibliographystyle{IEEEtran}
\bibliography{sample-base}

\end{document}